\documentclass[11pt,a4paper]{article}
\usepackage[hyperref]{acl2020}
\usepackage{times}
\usepackage{latexsym}

\usepackage{multirow}
\usepackage{comment}
\usepackage{listings}

\usepackage{microtype}
\usepackage{todonotes}
\usepackage{hyperref}
\usepackage{subcaption}

\aclfinalcopy 

\title{Transformer-based Methods for Recognizing Ultra Fine-grained Entities (RUFES)}

\author{Emanuela Boros \and
  Antoine Doucet \\
  L3i, University of La Rochelle, \\
  F-17000, La Rochelle, France \\
  \texttt{\{emanuela.boros,antoine.doucet\}@univ-lr.fr} \\}

\date{}

\begin{document}
\maketitle
\begin{abstract}
This paper summarizes the participation of the \textit{Laboratoire Informatique, Image et Interaction} (L3i laboratory) of the University of La Rochelle in the Recognizing Ultra Fine-grained Entities (RUFES) track\footnote{\url{https://tac.nist.gov/2020/KBP/RUFES/}} within the Text Analysis Conference (TAC) series of evaluation workshops. Our participation relies on two neural-based models, one based on a pre-trained and fine-tuned language model with a stack of Transformer layers for fine-grained entity extraction and one out-of-the-box model for within-document entity coreference. We observe that our approach has great potential in increasing the performance of fine-grained entity recognition. Thus, the future work envisioned is to enhance the ability of the models following additional experiments and a deeper analysis of the results. 
\end{abstract}

\section{Introduction}

Fine-grained entity recognition aims at labeling entity mentions in context with one or more specific types organized in a hierarchy (e.g., \textit{Photographer} is from a \textit{Artist} that in turn, is a subtype of \textit{PER}\footnote{\textit{PER} refers to the entity type \textit{Person}.}). The need for a wider variety of fine-grained entities  (e.g., technical terms, lawsuits, disease, crisis, biomedical entities) \cite{jioverview} can support the development of real-world applications that combine several information sources that include both text sources and knowledge bases; e.g., question answering \cite{lin2019attentive}, relation extraction systems \cite{yao2010collective} that need access to knowledge bases i.e. gazetteers, or named entity recognition systems \cite{ehrmann2020introducing,ehrmann2020impresso} for supporting more accurate entity linking in historical documents.

In the context of the Recognizing Ultra Fine-grained Entities (RUFES) track\footnote{\url{https://tac.nist.gov/2020/KBP/RUFES/}} of the Text Analysis Conference (TAC) series of evaluation workshops, the main task consisted in automatically identifying fine-grained entities as clusters of names, nominals, and/or pronominal mentions, and classifying them into one or more of the types defined in a detailed ontology developed by NIST\footnote{\url{https://www.nist.gov/}}. This year, the track focused on document-level entity discovery and only on  English source documents.  

The track had two phases: a preliminary phase where the data is provided along with a limited annotated set of samples ($50$ documents), and a second phase during which human feedback was provided for the preliminary submissions based on a  user model of how analysts might interact with the systems. The final results include this feedback.

This paper presents the participation of the \textit{Laboratoire Informatique, Image et Interaction} (L3i laboratory) at the University of La Rochelle at TAC KBP RUFES 2020. We applied our recent proposed model for coarse-grained and fine-grained named entity recognition  \cite{borocs2020alleviating,boros2020robust} and we used an out-of-the-box neural-based entity coreference model for detecting the mentions that refer to the same entity. 

The remaining of the paper is organized as follows: we detail the data pre-processing step and our proposed methods in Section \ref{section:methods}. Section \ref{section:results} presents the results and Section \ref{section:conclusions} finalizes the paper with several conclusions and envisioned future work.

\section{Methods}\label{section:methods}

We separated RUFES in two sub-tasks:
\begin{itemize}
    \item Entity extraction: the detection and the classification of fine-grained entity types including the named,  nominal, and pronominal mentions for each mention (labeled as NAM, NOM, and PRO, respectively);
    \item Within-document entity coreference resolution: the detection of the referential mentions in a document that point to the same entity.
\end{itemize}

\subsection{Data Pre-processing}
The KBP 2020 RUFES dataset provided by the organizers consisted of the development source documents and evaluation source documents drawn from a collection of Washington Post news articles. 

The development source corpus and the evaluation source corpus each comprised approximately $100,000$ articles from which $50$ documents were annotated for the development set with entity types from an ontology that contains approximately $200$ fine-grained entity types and that followed the same three-level x.y.z hierarchy as in the TAC-KBP 2019 EDL  track \cite{jioverview}.


The provided data was organized into two formats: \texttt{./rsd/}: ``raw source data'' (rsd) plain text form of the new article; and \texttt{./ltf/}: ``logical text format'' (ltf) derived from the {rsd} version.

For the data pre-processing, we used the \texttt{ltf.xml} files that each contained a fully segmented and tokenized version of the corresponding {rsd} file\footnote{Segments (paragraphs) and the tokens (words) are marked off by XML tags (SEG and TOKEN), with ``id'' attributes (which are only unique within a given XML file) and character offset attributes relative to the corresponding \texttt{rsd.txt} file.}. Next, we converted the data in IOB\footnote{\url{https://en.wikipedia.org/wiki/Inside\%E2\%80\%93outside\%E2\%80\%93beginning_(tagging)}} format as shown in the following example: 

\begin{figure}[ht]
  \centering
  \includegraphics[width=1.\linewidth]{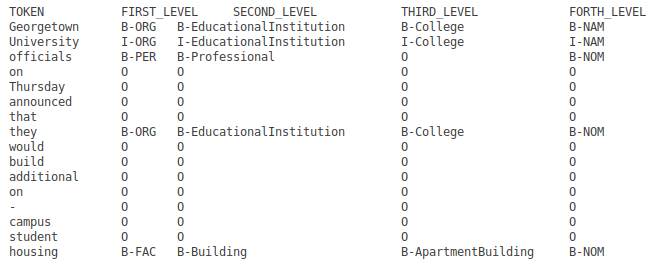}
  \caption{Data formatting example for the KBP 2020 RUFES dataset.}
  \label{fig:model_part_ner}
\end{figure}

\subsection{Entity Extraction}

\begin{figure}
\centering

  \centering
  \includegraphics[width=.8\linewidth]{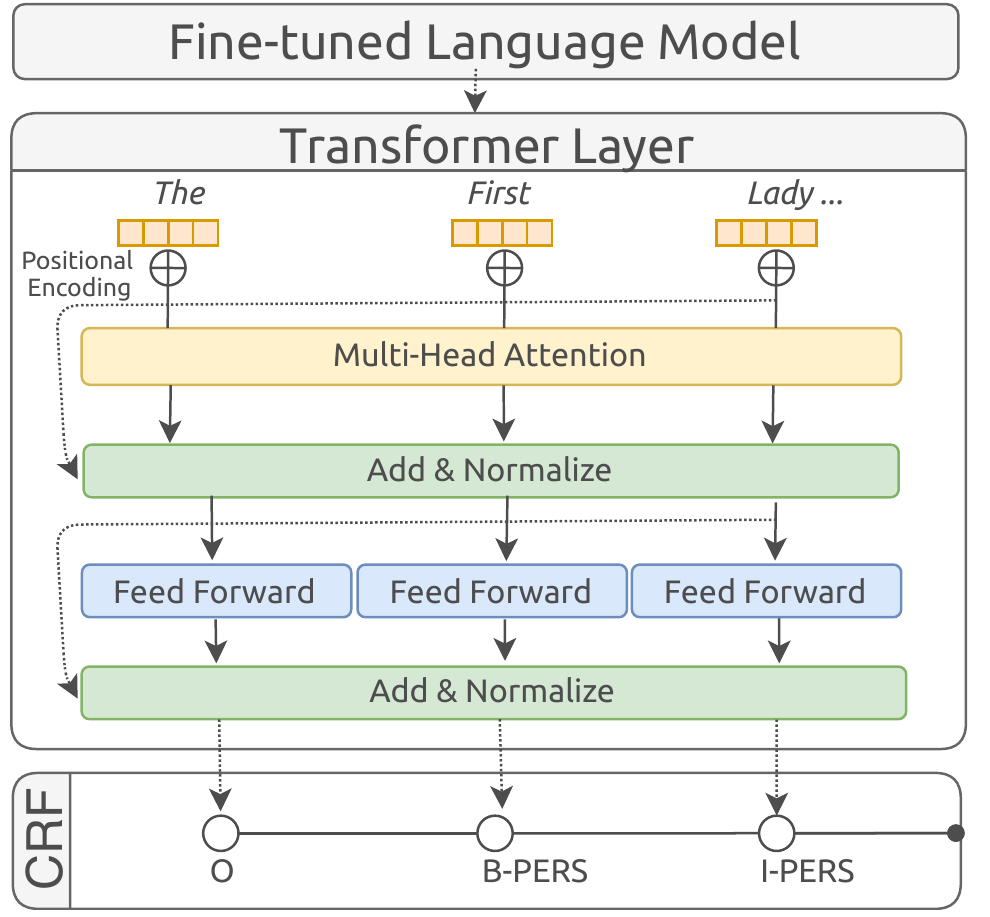}
  \caption{Detailed model proposed for entity extraction in \newcite{borocs2020alleviating,boros2020robust}.}
  \label{fig:model_part_ner}
\end{figure}

Due to the complexity and characteristics of the fine-grained sub-task, we made use of our recently proposed model for coarse-grained and fine-grained named entity recognition \cite{borocs2020alleviating,boros2020robust} that consists in a hierarchical, multitask learning approach, with a fine-tuned encoder based on \textit{Bidirectional Encoder Representations from Transformers} (BERT) \cite{devlin2018bert}. 

This model includes the use of a stack of Transformer \cite{vaswani2017attention} blocks on top of the BERT encoder. The multitask prediction layer consists of separate conditional random field (CRF) layers. The architecture of the model is presented in Figure \ref{fig:model_part_ner}.


The added layers are a stack of Transformer blocks (encoders). As proposed by \newcite{vaswani2017attention}, this model is a deep learning architecture based on multi-head attention mechanisms with \textit{sinusoidal position embeddings} \footnote{In our implementation, we used \textit{learned absolute positional embeddings} \cite{gehring2017convolutional} instead, as suggested by \newcite{Wolf2019HuggingFacesTS}. \newcite{vaswani2017attention} found that both versions produced nearly identical results.}. It is composed of a stack of identical layers. Each layer has two sub-layers. The first layer is a multi-head self-attention mechanism, while the second one is a simple, position-wise fully connected feed-forward network. A residual connection is around each of the two sub-layers, followed by layer normalization. 

We decided to add a stack of Transformer layers due to the assumption that additional hyperparameters can increase the ability of the architecture to better model long-range contexts and alleviate the number of spurious predicted entities, as observed in \newcite{borocs2020alleviating}.

Because of multitask learning, this method has a label independence assumption, which is not valid for fine-grained entity typing. For example, if the model is confident at predicting the type \textit{Photographer}, it should promote its parents of type \textit{Artist} and \textit{PER}, but discourage entity types as in \textit{ORG} and its descendant types (i.e. \textit{Association}, \textit{CommercialOrganization}). In order to capture interdependencies between types, we post-process the predictions by checking them again against the ontology terms, and by offering more confidence to the last predicted entity subtype. For example, if we take the subtype \textit{ProvinceState} with its parent \textit{GPE}, and if ``Illinois'' was recognized as \textit{LOC.ProvinceState}, we lookup \textit{ProvinceState} in the ontology and we automatically correct the prediction to \textit{GPE.ProvinceState}.

\paragraph{Parameters}

For the pre-trained BERT encoder, we used the \texttt{bert-large-cased} model. We added two Transformer layers, with the hidden size of $128$ and $12$ number of self-attention heads\footnote{The parameters correspond to the best configuration reported by \newcite{borocs2020alleviating}.}. We trained for $10$ epochs, with Adam optimizer with weight decay, $\mathrm{2\times10}^{-5}$ learning rate and a mini-batch of dimension $8$.

\subsection{Within-document Entity Coreference Resolution}

For detecting the referential mentions in a document that point to the same entity, we used an out-of-the-box tool, NeuralCoref\footnote{\url{https://github.com/huggingface/neuralcoref}}, which is a pipeline extension for spaCy 2.1+\footnote{\url{https://spacy.io/}} \cite{spacy2} that annotates and resolves coreference clusters using a neural-based method. Due to a lack of time and resources, we did not re-train this model on the KBP 2020 RUFES dataset.
The model was previously trained on OntoNotes 5.0 dataset\footnote{\url{https://www.gabormelli.com/RKB/OntoNotes_Corpus}}.

NeuralCoref has two sub-modules:
\begin{itemize}
    \item a rule-based mentions detection module which uses spaCy's tagger, parser and entity annotations to identify a set of potential coreference mentions;
    \item a feed-forward neural-network which compute a coreference score for each pair of potential mentions. This scoring system is an adaptation of \newcite{clark2016deep,clark2016improving}.
\end{itemize}

We applied this model in a within-document context, with the default parameters.

 


\subsection{Rule-based Feedback Inclusion}


In the second phase of the track, after receiving the feedback, we inspected the most frequent types of error produced by our methods. 

A majority of the mistakes, around $46$\% out of $400$ (the first $40$ errors detected in ten random documents were reported) were system mention-level entity types that do not exactly match the gold mention-level entity types (including the level of granularity). Around $11$\% were extraneous mentions (a system mention span does not exactly match or overlap with any gold mention span), $5$\% of wrong extents (a system mention span and gold mention span overlap but have different extents), $12$\% missing mentions (a gold mention span does not exactly match any system mention span), and around $11$\% had the wrong entity coreference, either missing, incorrect or spurious (problems in linking a system mention to a mention of the same entity that occurred earlier in the document).

Thus, we focused on the feedback related to the detection of the wrong type of entity ($46$\% out of all errors), and we integrated this feedback in a rule-based manner by automatically creating a set of rules to change the predictions accordingly. One common mistake produced by our system was related to entities that had one of the ontology terms included in the entity; For instance, ``Norovirus'' was recognized as GPE (geopolitical entity) instead of Pathogen.Virus. Therefore, for every entity that included a fine-grained ontology type i.e. ``Airport'', ``Hospital'', ``Highway'', a rule was created to change the predictions into the correct types.



\section{Results}\label{section:results}

\begin{table*}[ht]
\centering
\begin{tabular}{p{2.9cm}p{1.8cm}p{1.8cm}p{1.6cm}p{1.5cm}p{1.5cm}p{1.7cm}}
\hline 

\textbf{Submission} & \textbf{strong mention match} & \textbf{strong typed mention match} & \textbf{mention ceaf} &  \textbf{typed mention ceaf} & \textbf{entity ceaf} & \textbf{fine grain typing} \\ \hline

1-first-rufes & \bf 0.868 &	0.745 & 0.552 &	0.503 & 0.551 & 0.3188 \\
2-first-rufes & \bf 0.868 &	0.745 & 0.578 & 0.503  & 0.567 & 0.3188 \\ \hline
1-feedback-rufes & \bf 0.868 & 0.745  & 0.578 & 0.504  & 0.567 & 0.3204 \\ 
2-feedback-rufes & \bf 0.868 & 0.745 & 0.578 & 0.504  & 0.567 & 0.3239 \\
Median & 0.805 & -- & -- &  -- & 0.578 & 0.2313 \\
Maximum & \textbf{{0.868}} & -- & -- & -- & 0.689 & \textbf{{0.4162}} \\

\hline
\end{tabular}
\caption{\label{table:results} The scoring results (F-score) for RUFES 2020 evaluation for all our submissions. Median and Maximum scores are computed on the best-performing submission from each participant, as shared by RUFES organizers.}
\end{table*}


In the initial phase, we submitted two runs named \textit{1-first-rufes} and \textit{2-first-rufes}. In the second phase, after receiving feedback for a sample of our previous predictions, we submitted another two runs referred to as \textit{1-feedback-rufes} and \textit{2-feedback-rufes}. 

The metrics reported in Table \ref{table:results} are implemented by neleval\footnote{More details about the TAC evaluation tool can be found here \url{https://neleval.readthedocs.io/en/latest/}.}. For the entity extraction evaluation, we report the \textit{strong\_mention\_match} (where an entity span must match a gold span exactly to be counted as correct) and the \textit{strong\_typed\_mention\_match} (that additionally requires the correct entity type). For the entity coreference evaluation \cite{pradhan2014scoring}, we report the \textit{typed\_mention\_ceaf}, the \textit{entity\_ceaf}, and \textit{mention\_ceaf} metrics. We also report the RUFES final score, \textit{fine\_grain\_typing}.

From the scoring results for our submissions reported in the Table \ref{table:results}, we notice that the impact of the ruled-based feedback inclusion is rather insignificant. The slight differences in the scores when comparing \textit{1-first-rufes} and the other submissions regarding the \textit{entity\_ceaf} and \textit{mention\_ceaf} metrics are due to the fact that we did not approach the entity coreference for our first submission.





\section{Conclusions}\label{section:conclusions}

This paper described our approach for the 2020 TAC RUFES task that implied fine-grained entity recognition and within-document entity coreference. We presented our proposed models, and we reported the results obtained in the context of the track. In future work, we will focus on enhancing the ability of the models following additional experiments by refining the entity extraction architecture in order to be able to take into consideration the inter-dependencies between entity types. Regarding the entity coreference model, we could explore a further fine-tuning of the out-of-the-box model on the KBP RUFES annotated documents. Moreover, a deeper and more qualitative analysis of the types of errors is intended. 

\section*{Acknowledgments}

This work has been supported by the European Union's Horizon 2020 research and innovation program under grants 770299 (NewsEye) and 825153 (Embeddia).

\bibliography{acl2020}
\bibliographystyle{acl_natbib}

\end{document}